%% file: main.tex
\documentclass[conference]{acmsiggraph}

\usepackage{balance}  
\usepackage{graphics} 
\usepackage{times}    
\usepackage{url}      
\usepackage[activate]{pdfcprot}
\usepackage[usenames,dvipsnames]{xcolor}
\usepackage{tabularx}
\usepackage{booktabs}
\usepackage{subcaption}

\newcommand{\owntitle}{Labeled pupils in the wild: A dataset for studying pupil detection in unconstrained environments}
\newcommand{\ownkeywords}{Pupil detection; Head-mounted eye tracking; High-speed; High-quality}

\newcolumntype{P}[1]{>{\raggedright\arraybackslash}m{#1}}%
\newcolumntype{C}[1]{>{\centering\arraybackslash}m{#1}}%
\newcolumntype{R}[1]{>{\raggedleft\arraybackslash}m{#1}}%

\title{\owntitle}

\author{
Marc Tonsen
\and
Xucong Zhang
\and
Yusuke Sugano
\and
Andreas Bulling\\
\and
Perceptual User Interfaces Group\\
Max Planck Institute for Informatics, Saarbr\"ucken, Germany\\
{\tt\small \{tonsen,xczhang,sugano,bulling\}@mpi-inf.mpg.de}
}
\pdfauthor{Marc Tonsen, Xucong Zhang, Yusuke Sugano, Andreas Bulling}


\begin{document}

\maketitle

\begin{abstract}
\input{00_abstract.tex}
\end{abstract}

\begin{CRcatlist}
  \CRcat{I.4.9}{Image Processing and Computer Vision}{Applications}{};
\end{CRcatlist}

\keywordlist

\input{01_intro.tex}

\input{03_method.tex}

\input{04_evaluation.tex}

\input{06_discussion.tex}

\input{07_conclusion.tex}

\bibliographystyle{acmsiggraph}
\bibliography{main}
\end{document}

%% file: 00_abstract.tex

We present labelled pupils in the wild (LPW), a novel dataset of 66 high-quality, high-speed eye region videos for the development and evaluation of pupil detection algorithms.
The videos in our dataset were recorded from 22 participants in everyday locations at about 95 FPS using a state-of-the-art dark-pupil head-mounted eye tracker.
They cover people with different ethnicities, a diverse set of everyday indoor and outdoor illumination environments, as well as natural gaze direction distributions.
The dataset also includes participants wearing glasses, contact lenses, as well as make-up.
We benchmark five state-of-the-art pupil detection algorithms on our dataset with respect to robustness and accuracy.
We further study the influence of image resolution, vision aids, as well as recording location (indoor, outdoor) on pupil detection performance.
Our evaluations provide valuable insights into the general pupil detection problem and allow us to identify key challenges for robust pupil detection on head-mounted eye trackers. 

%% file: 01_intro.tex

\section{Introduction}

\begin{figure}
    \begin{subfigure}[t]{0.24\columnwidth}
        \includegraphics[width=\columnwidth]{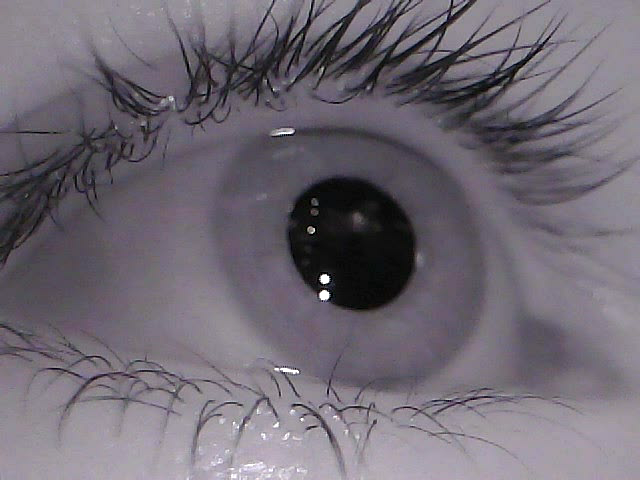}
        \caption{}
    \end{subfigure}
    \begin{subfigure}[t]{0.24\columnwidth}
        \includegraphics[width=\columnwidth]{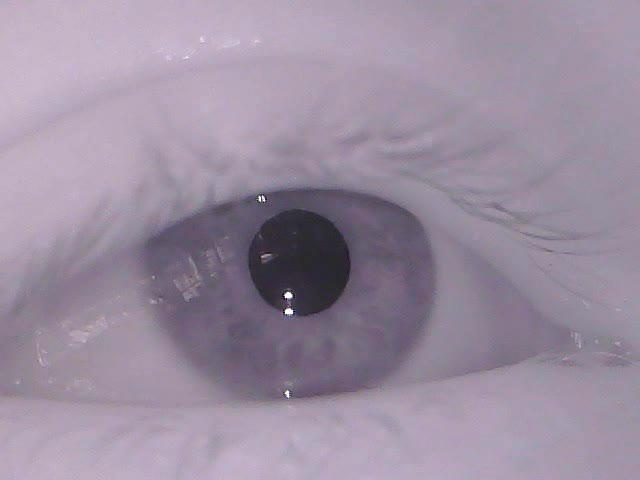}
        \caption{}
    \end{subfigure}
    \begin{subfigure}[t]{0.24\columnwidth}
        \includegraphics[width=\columnwidth]{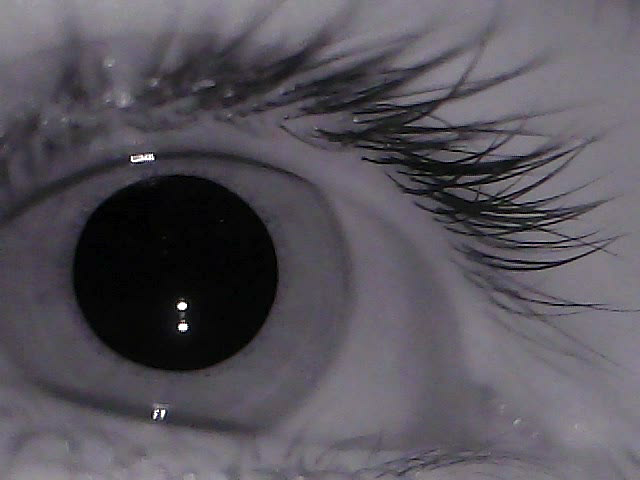}
        \caption{}
    \end{subfigure}
    \begin{subfigure}[t]{0.24\columnwidth}
        \includegraphics[width=\columnwidth]{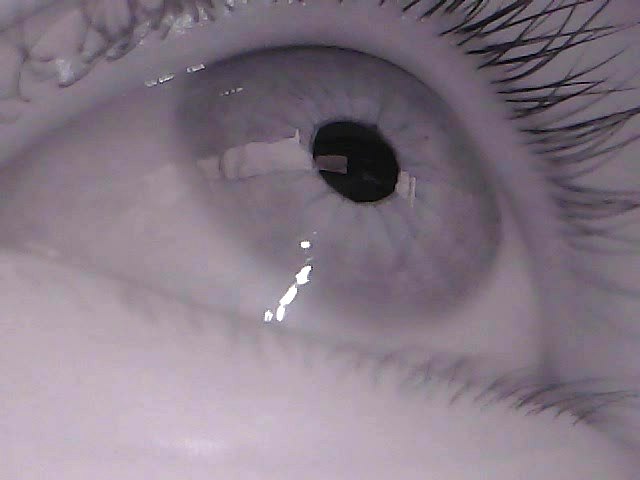}
        \caption{}
    \end{subfigure}
    
    \begin{subfigure}[t]{0.24\columnwidth}
        \includegraphics[width=\columnwidth]{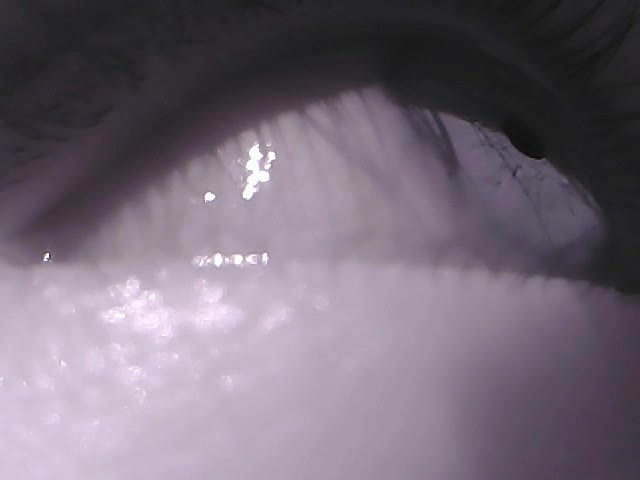}
        \caption{}
    \end{subfigure}
    \begin{subfigure}[t]{0.24\columnwidth}
        \includegraphics[width=\columnwidth]{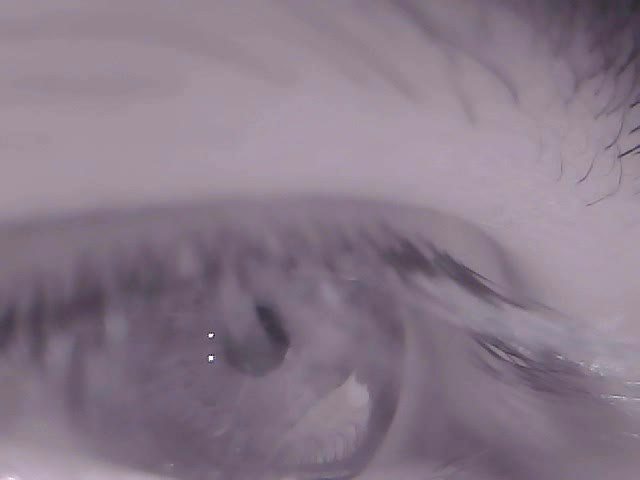}
        \caption{}
    \end{subfigure}
    \begin{subfigure}[t]{0.24\columnwidth}
        \includegraphics[width=\columnwidth]{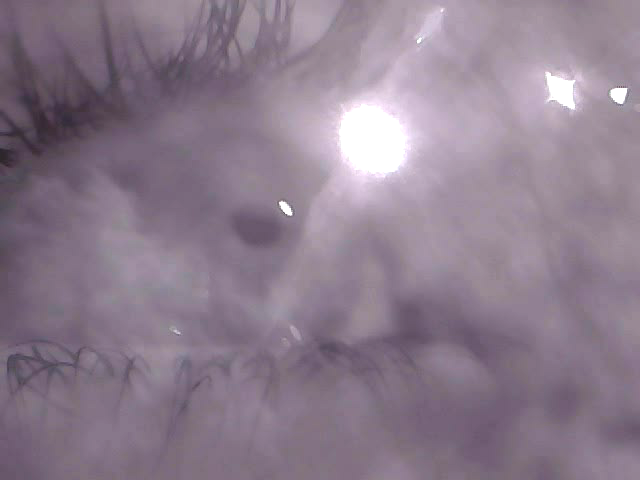}
        \caption{}
    \end{subfigure}
    \begin{subfigure}[t]{0.24\columnwidth}
        \includegraphics[width=\columnwidth]{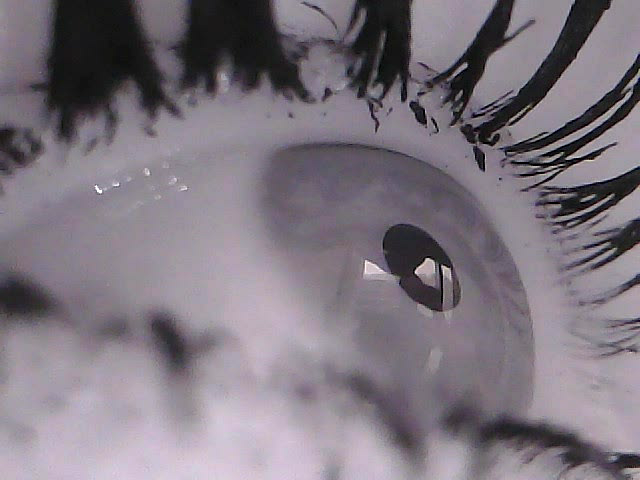}
        \caption{}
    \end{subfigure}
    
    \begin{subfigure}[t]{0.24\columnwidth}
        \includegraphics[width=\columnwidth]{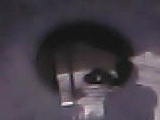}
        \caption{}
    \end{subfigure}
    \begin{subfigure}[t]{0.24\columnwidth}
        \includegraphics[width=\columnwidth]{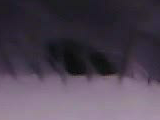}
        \caption{}
    \end{subfigure}
    \begin{subfigure}[t]{0.24\columnwidth}
        \includegraphics[width=\columnwidth]{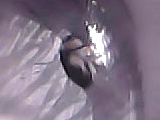}
        \caption{}
    \end{subfigure}
    \begin{subfigure}[t]{0.24\columnwidth}
        \includegraphics[width=\columnwidth]{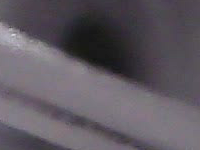}
        \caption{}
    \end{subfigure}
    \vspace{2mm}
    \caption{Example images of variability in our dataset. The first row (a) (b) (c) and (d) shows different eye appearances. The second row shows the most difficult cases according to our evaluation: (e) strong shade, (f) eyelid occlusion, (g) reflection on glasses, (h) strong makeup. The third row shows crops around pupil under challenging conditions: (i) reflection on the pupil, (j) self occluded, (k) strong sunlight and shade, (l) occlusion by glasses.}
    \label{fig:teaser}
\end{figure}

Pupil detection is a core component of shape-based gaze estimation systems and therefore well-established as a research topic in eye tracking~\cite{hansen2010eye}.
Robust and accurate pupil detection is challenging, particularly in eye images recorded using head-mounted eye trackers.
These systems are used in mobile everyday settings and eye images can therefore become subject to significant influences by changes in ambient light, corneal reflections, pupil occlusions, and shadows (see~\autoref{fig:teaser}).
Despite considerable advances, we argue that methods for pupil detection on head-mounted eye trackers lack behind.
When analysing current benchmark datasets, we identified two main limiting factors.

First, several existing datasets were recorded using remote cameras and only consist of monocular RGB images (see~\cite{jesorsky2001robust} for an example).
Images recorded under these conditions are significantly different from the close-up infrared eye region images recorded on head-mounted eye trackers.
Second, the few datasets for head-mounted pupil detection that are publicly available are either limited in size, were recorded in controlled laboratory settings and therefore do not cover realistic day-to-day usage scenarios -- that, for example, also include transitions of users between indoor and outdoor environments -- or only contain low-quality eye images (see \autoref{tab:other_datasets} for a comparison).

The dataset presented in \cite{swirski_dataset} includes 600 high-quality close-up eye images and manual ground truth annotations of the pupil center.
While this dataset is a good starting point to evaluate pupil detection algorithms, it is limited in that it only contains eye images of two participants and was collected in the laboratory with controlled lighting conditions.
A more recent dataset was introduced in~\cite{excuse}.
The dataset is significantly larger than the first dataset and images were recorded with a head-mounted eye tracker in uncontrolled environments, namely while driving and going shopping, but not in fully outdoor environments.

In this paper we therefore present \textit{labelled pupils in the wild} (LPW), a novel pupil detection dataset that aims to address these shortcomings.
More specifically, we present a dataset of 66 high-quality eye region videos that were recorded from 22 participants using a state-of-the-art dark-pupil head-mounted eye tracker.
Each video in the dataset consists of about 2,000 frames with a resolution of 640x480 pixels and was recorded at about 95 FPS, resulting in a total of 130,856 video frames.
The dataset is one order of magnitude larger than existing ones and 
covers a wide variety of realistic indoor and outdoor illumination conditions, include participants wearing glasses and eye make-up, as well as cover different ethnicities with variable skin tones, eye colours, and face shapes.
All videos were manually ground-truth annotated with accurate pupil ellipse and center positions.
We further evaluate several state-of-the-art pupil detection algorithms on this challenging new dataset. 
Our evaluations provide valuable insights into the pupil detection problem setting and allow us to identify key challenges for pupil detection on head-mounted eye trackers.
The full dataset and ground truth annotations will be made publicly available upon acceptance.

%% file: 03_method.tex

\section{Labelled pupils in the wild (LPW) dataset}

\begin{table*}[ht]
\footnotesize
\centering
\begin{tabularx}{\textwidth}{c c c c c c c c r}
& \textbf{participants} & \textbf{sessions} & \textbf{images} & \textbf{camera angles} & \textbf{lighting conditions} & \textbf{ethnicities} & \textbf{resolution} & \textbf{FPS}\\ 
\hline 
\cite{swirski_dataset} & 2 & 4 & 600 & 4 & 1 & n.a. & 640x480 & static images\\ 
\hline 
\cite{excuse} & 17 & 17 & 38,401 & mostly frontal & $\leq$ 17 & n.a. & 384x288 & 25\\ 
\hline 
Ours & 22 & 66 & 130,856 & continuous & continuous & 5 & 640x480 & 95
\end{tabularx} 
\caption{Comparison of current publicly available datasets for pupil detection on head-mounted eye trackers.}
\label{tab:other_datasets}
\end{table*}

We designed a data collection procedure with two main goals in mind: 1) to record samples of participants under different conditions, i.e. different lighting conditions and eye camera positions, and 2) to have a large variability in appearance of participants, such as gender, ethnicity and use of vision aids. We took each participant to a different set of locations and recorded their eye movements while looking at a moving gaze target.

\def\arraystretch{1.3}%
\begin{table*}[ht]
\footnotesize
\centering
\begin{tabularx}{\textwidth}{C{1.4cm} *{11}{|C{1.0cm}}}
 & \textbf{P01 (m)} & \textbf{P02 (m)} &  \textbf{P03 (f)} &  \textbf{P04 (m)} &  \textbf{P05 (f)} &  \textbf{P06 (n)} &  \textbf{P07 (m)} &  \textbf{P08 (m)} &  \textbf{P09 (m)} &  \textbf{P10 (f)} &  \textbf{P11 (m)} \\
 \toprule
 Nationality & Iranian & German & Iranian & Indian & German & Indian & Indian & Pakistani & German & Indian & Pakistani \\
 \hline 
 Eye color & Brown & Blue & Brown & Brown & Brown & Black & Black & Brown & Blue-gray & Brown & Brown \\
 \hline
 Glasses & No & No & Yes & No & Yes & No & No & No & No & No & No \\
 \hline
 Video variability & 
  \begin{tabular}{@{}P{0.30cm} | P{0.15cm}}In & Out \\ \hline 2 & 1 \\ \hline Nat & Art \\ \hline 3 & 1 \\\end{tabular} &
  \begin{tabular}{@{}P{0.30cm} | P{0.15cm}}In & Out \\ \hline 2 & 1 \\ \hline Nat & Art \\ \hline 3 & 1 \\\end{tabular} &
  \begin{tabular}{@{}P{0.30cm} | P{0.15cm}}In & Out \\ \hline 2 & 1 \\ \hline Nat & Art \\ \hline 2 & 2 \\\end{tabular} &
  \begin{tabular}{@{}P{0.30cm} | P{0.15cm}}In & Out \\ \hline 2 & 1 \\ \hline Nat & Art \\ \hline 2 & 1 \\\end{tabular} &
  \begin{tabular}{@{}P{0.30cm} | P{0.15cm}}In & Out \\ \hline 1 & 2 \\ \hline Nat & Art \\ \hline 3 & 1 \\\end{tabular} &
  \begin{tabular}{@{}P{0.30cm} | P{0.15cm}}In & Out \\ \hline 2 & 1 \\ \hline Nat & Art \\ \hline 2 & 1 \\\end{tabular} &
  \begin{tabular}{@{}P{0.30cm} | P{0.15cm}}In & Out \\ \hline 1 & 2 \\ \hline Nat & Art \\ \hline 3 & 1 \\\end{tabular} &
  \begin{tabular}{@{}P{0.30cm} | P{0.15cm}}In & Out \\ \hline 2 & 1 \\ \hline Nat & Art \\ \hline 3 & 1 \\\end{tabular} &
  \begin{tabular}{@{}P{0.30cm} | P{0.15cm}}In & Out \\ \hline 2 & 1 \\ \hline Nat & Art \\ \hline 3 & 1 \\\end{tabular} &
  \begin{tabular}{@{}P{0.30cm} | P{0.15cm}}In & Out \\ \hline 2 & 1 \\ \hline Nat & Art \\ \hline 2 & 1 \\\end{tabular} & 
  \begin{tabular}{@{}P{0.30cm} | P{0.15cm}}In & Out \\ \hline 2 & 1 \\ \hline Nat & Art \\ \hline 1 & 2 \\\end{tabular} \\
\bottomrule
\end{tabularx}\\

\vspace*{0.3cm}

\begin{tabularx}{\textwidth}{C{1.4cm} *{11}{|C{1.0cm}}}
 &  \textbf{P12 (m)} &  \textbf{P13 (m)} &  \textbf{P14 (f)} &  \textbf{P15 (f)} &  \textbf{P16 (m)} &  \textbf{P17 (m)} &  \textbf{P18 (m)} &  \textbf{P19 (f)} &  \textbf{P20 (f)} &  \textbf{P21 (f)} &  \textbf{P22 (f)} \\
\toprule
 Nationality & Egyptian & Indian & Indian & German & German & Indian & Indian & Indian & Indian & Indian & German \\
 \hline
 Eye color & Brown & Black & Brown & Blue-gray & Green & Brown & Brown & Black & Black & Brown & Blue-gray\\ 
 \hline
 Glasses & No & No & No & No & Contact lenses & No & No & No & No & No & No \\
 \hline
 Video variability & 
  \begin{tabular}{@{}P{0.30cm} | P{0.15cm}}In & Out \\ \hline 2 & 1 \\ \hline Nat & Art \\ \hline 3 & 1 \\\end{tabular} &
  \begin{tabular}{@{}P{0.30cm} | P{0.15cm}}In & Out \\ \hline 2 & 1 \\ \hline Nat & Art \\ \hline 3 & 1 \\\end{tabular} &
  \begin{tabular}{@{}P{0.30cm} | P{0.15cm}}In & Out \\ \hline 2 & 1 \\ \hline Nat & Art \\ \hline 2 & 1 \\\end{tabular} &
  \begin{tabular}{@{}P{0.30cm} | P{0.15cm}}In & Out \\ \hline 2 & 1 \\ \hline Nat & Art \\ \hline 3 & 1 \\\end{tabular} &
  \begin{tabular}{@{}P{0.30cm} | P{0.15cm}}In & Out \\ \hline 2 & 1 \\ \hline Nat & Art \\ \hline 3 & 1 \\\end{tabular} &
  \begin{tabular}{@{}P{0.30cm} | P{0.15cm}}In & Out \\ \hline 2 & 1 \\ \hline Nat & Art \\ \hline 3 & 1 \\\end{tabular} &
  \begin{tabular}{@{}P{0.30cm} | P{0.15cm}}In & Out \\ \hline 2 & 1 \\ \hline Nat & Art \\ \hline 2 & 1 \\\end{tabular} &
  \begin{tabular}{@{}P{0.30cm} | P{0.15cm}}In & Out \\ \hline 2 & 1 \\ \hline Nat & Art \\ \hline 3 & 1 \\\end{tabular} &
  \begin{tabular}{@{}P{0.30cm} | P{0.15cm}}In & Out \\ \hline 2 & 1 \\ \hline Nat & Art \\ \hline 2 & 1 \\\end{tabular} &
  \begin{tabular}{@{}P{0.30cm} | P{0.15cm}}In & Out \\ \hline 2 & 1 \\ \hline Nat & Art \\ \hline 2 & 2 \\\end{tabular} & 
  \begin{tabular}{@{}P{0.30cm} | P{0.15cm}}In & Out \\ \hline 3 & 0 \\ \hline Nat & Art \\ \hline 2 & 1 \\\end{tabular} \\
\bottomrule
\end{tabularx}
\caption{Characteristics of the LPW dataset. The gender of participants has been indicated as female (f) and male (m). The variability of videos is represented as indoor (In) and outdoor (Out), with natural (Nat) and artificial (Art) light.}
\label{tab:dataset_characters}
\end{table*}

\begin{figure}
	\begin{minipage}{0.677\columnwidth}
		\begin{subfigure}[t]{\columnwidth}
        	\includegraphics[width=\columnwidth]{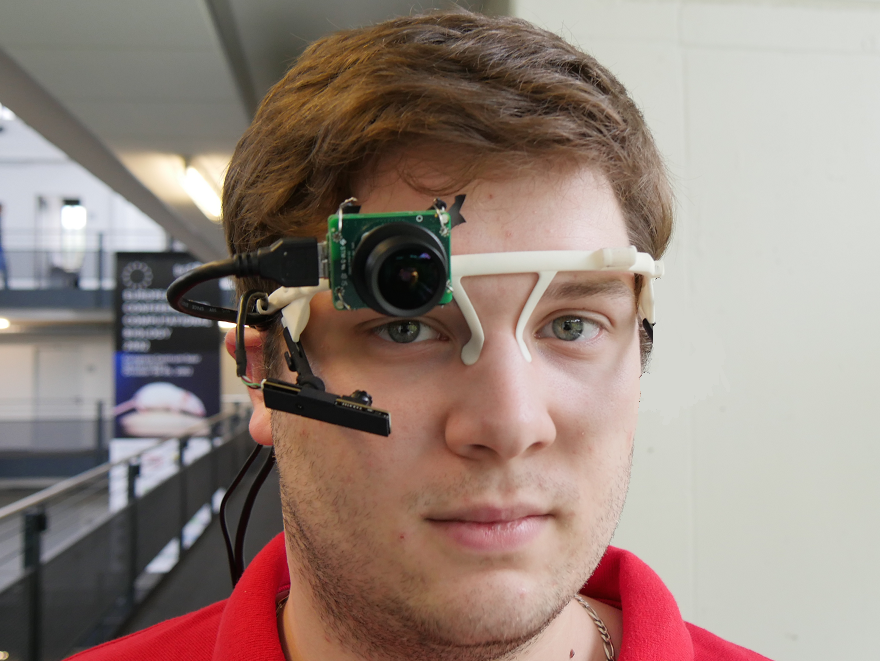}
        	\caption{}
        	\label{fig:pupil_tracker}
    	\end{subfigure}
	\end{minipage}
	\begin{minipage}{0.313\columnwidth}
		\begin{subfigure}[t]{\columnwidth}
        	\includegraphics[width=\columnwidth]{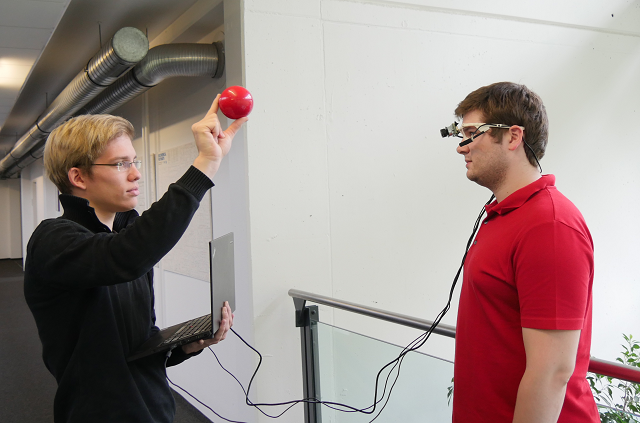}
        	\caption{}
        	\label{fig:system_setting}
    	\end{subfigure}
    
    	\begin{subfigure}[t]{\columnwidth}
        	\includegraphics[width=\columnwidth]{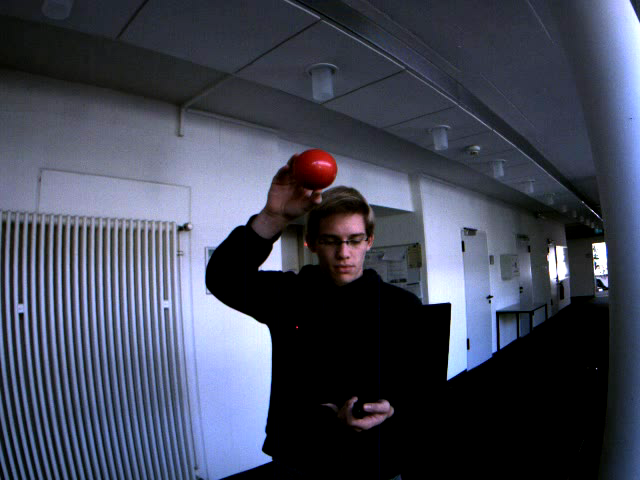}
        	\caption{}
        	\label{fig:scene_img}
    	\end{subfigure}
	\end{minipage}

\caption{Data collection setting. (a) The high frame rate eye and scene cameras. (b) Participants move their eye by looking at the red ball. (c) The image captured by the scene camera.}
\label{fig:system}
\end{figure}

\subsection*{Participants}

Detailed information about our participants can be found in Table~\ref{tab:dataset_characters}. 
We recruited 22 participants including 9 female through university mailing lists and personal communication. 
Among them are five different ethnicities: 11 Indian, 6 German, 2 Pakistani, 2 Iranian, and 1 Egyptian. In total we had five different eye colors: 12 brown, 5 black, 3 blue-gray, 1 blue-green, 1 green. Also 5 people had impaired vision, 2 wore glasses and 1 wore contact lenses. Strong eye make-up was worn by 1 person (with participant ID 22).

\subsection*{Apparatus}

The eye tracker used for the recording was a high-speed Pupil Pro head-mounted eye tracker that record eye videos with 120 Hz~\cite{pupil}.
In order to capture high frame rate scene videos, we replaced the original scene camera with a PointGrey Chameleon3 USB3.0 camera recording at up to 149 Hz. The hardware set up is shown in Figure~\ref{fig:pupil_tracker} and Figure~\ref{fig:system_setting}.
It allowed us to record all videos with 95 FPS, 
which is a speed at which even fast eye movements last through several frames. 

\subsection*{Procedure}
As shown in Figure~\ref{fig:system_setting}, the participants were instructed to look at a moving red ball as a fixation target during the data collection.
The position of the red ball in the visual field of the participant is shown in Figure~\ref{fig:scene_img} with an image captured by the scene camera.

In order to cover as many different conditions as possible, we randomly picked the recording locations 
in and around of several buildings. Each location was not chosen more than once during the whole recording of all participants.
34.3\% of the recordings were done outdoors, in 84.7\% natural light was present and in 33.6\% artificial light was present.
Besides locations, we have also tweaked the angle of the eye cameras such that the dataset contains a wide range of camera angles from frontal views to highly off-axis angles.
This is done by either asking the participant to take the tracker off and put it back on, or manually moving the camera.
With each of the 22 participant we recorded three videos with around 20 seconds length, yielding 130,856 images overall.
Participants could keep their glasses and contact lenses on during the recording.

\subsection*{Ground truth annotation}
We used different methods for annotation.
In many easy cases such as some indoor recordings, the pupil area has a clear boundary and no strong reflections inside. We annotated these frames by manually selecting 1 or 2 points inside the pupil area, using them as seed points to find the largest connected area with similar intensity values. The pupil center is defined as the centroid of this area.

Some recordings have a clear scene video but strong reflections/noise in the eye video, such as outdoor recordings under strong sunlight. 
In those cases, we tracked the fixation target (red ball) in the scene videos and manually annotated part of the eye pupil positions in the eye videos.
From this calibration data we computed a mapping function from target positions to pupil positions. 
In addition, we examined the annotated videos again to find wrong annotations, and corrected them by selecting 5 or more points on the pupil boundary and fitting an ellipse to them. The center of the ellipse was used as a refined pupil center position.

%% file: 04_evaluation.tex

\section{Results}
To evaluate the difficulty and challenges contained in our dataset, we have analysed the performance of five state-of-the art pupil detection algorithms. {\em Pupil Labs}~\cite{pupil} is the algorithm used in the Pupil Pro eye tracker. 
{\em Swirski}~\cite{swirski_dataset} and {\em ExCuSe}~\cite{excuse} are taken as examples of the state-of-the-art algorithms. 
{\em Isophote}~\cite{valenti2012accurate} and {\em Gradient}~\cite{timm2011accurate} are two simple algorithms designed for the iris shape fitting task on low-resolution remote eye images. 
In the following sections we examine several performance values and highlight key challenges in our dataset.
We ran the evaluations on a Linux system desktop with an Intel E5800 CPU 3.16GHz processor and 8GB memory. The average processing speed of each algorithm was: {\em Isophote} 225.59 fps, {\em Pupil Labs} 45.09 fps,  {\em Gradient} 43.52 fps, {\em Swirski} 5.44 fps, {\em ExCuSe} 1.90 fps.

\subsubsection*{Accuracy and Robustness}
\begin{figure}[t]
\centering
\includegraphics[width = \columnwidth]{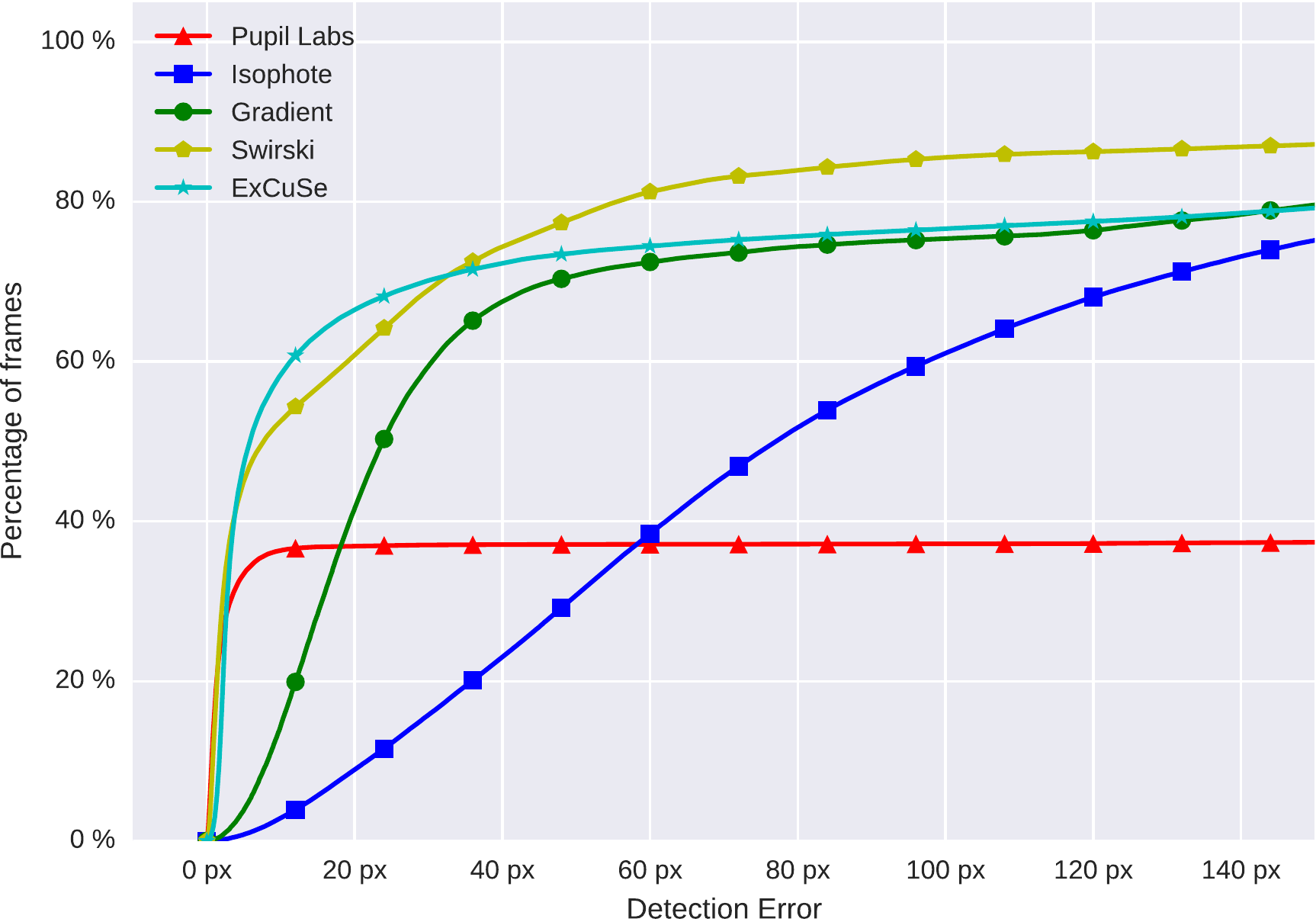}
\caption{Cumulative error distribution of each algorithm on the entire dataset. The x-axis describes the error in pixels, while the y-axis describes the percentage of detections that achieved an error smaller or equal to the corresponding x-value.}
\label{fig:cumulative_error_dist}
\end{figure}

Figure \ref{fig:cumulative_error_dist} shows the cumulative error distribution of all algorithms on the entire dataset. One can see that {\em Pupil Labs}, {\em Swirski} and {\em ExCuSe} all return very good results in roughly 30\% of all cases with less then 5px error; however their performances fall off quickly. It is worth mentioning that ExCuSe falls off last. The {\em Gradient} detector follows a similar curve but shifted to the right, indicating a higher error on average. The {\em Isophote} detector's curve rises the least steep indicating the highest error on average. {\em Pupil Labs} stands out by cutting off very early. While giving fairly accurate results in almost 40\% of all cases, it completely fails in the other 60\%. {\em ExCuSe}, {\em Swirski} and the {\em Gradient} detector return reasonable results with an error of roughly 40px in about 70\% of all cases, indicating a higher robustness in comparison to {\em Pupil Labs}. 

Overall there is no satisfying performance on the dataset yet for gaze estimation. This indicates the difficulty of our dataset, i.e., pupil detection in the wild is still challenging for current methods. According to our observations, the hardest samples are mainly cases of strong shadows, eyelid occlusions, reflections from glasses and strong make-up (see also Figure \ref{fig:teaser} (e), (f), (g) and (h)).

\subsubsection*{Indoor vs Outdoor} 
\begin{figure*}[t]
\centering
	\begin{subfigure}[t]{0.33\textwidth}
    	\includegraphics[width=\columnwidth]{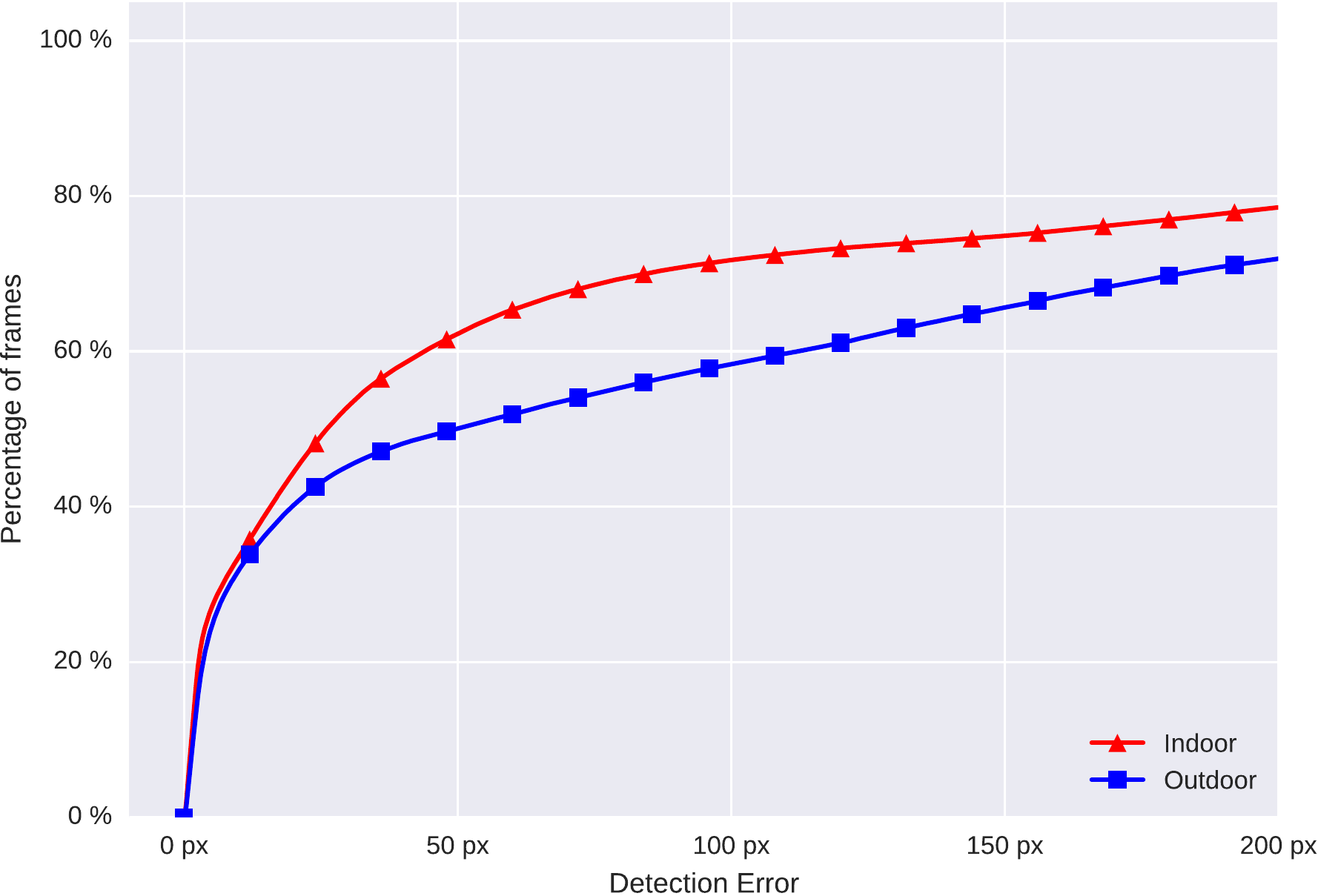}
    	\caption{Indoor/outdoor evaluation}
        \label{fig:in_out_door_eva}
    \end{subfigure}
    \begin{subfigure}[t]{0.33\textwidth}
    	\includegraphics[width=\columnwidth]{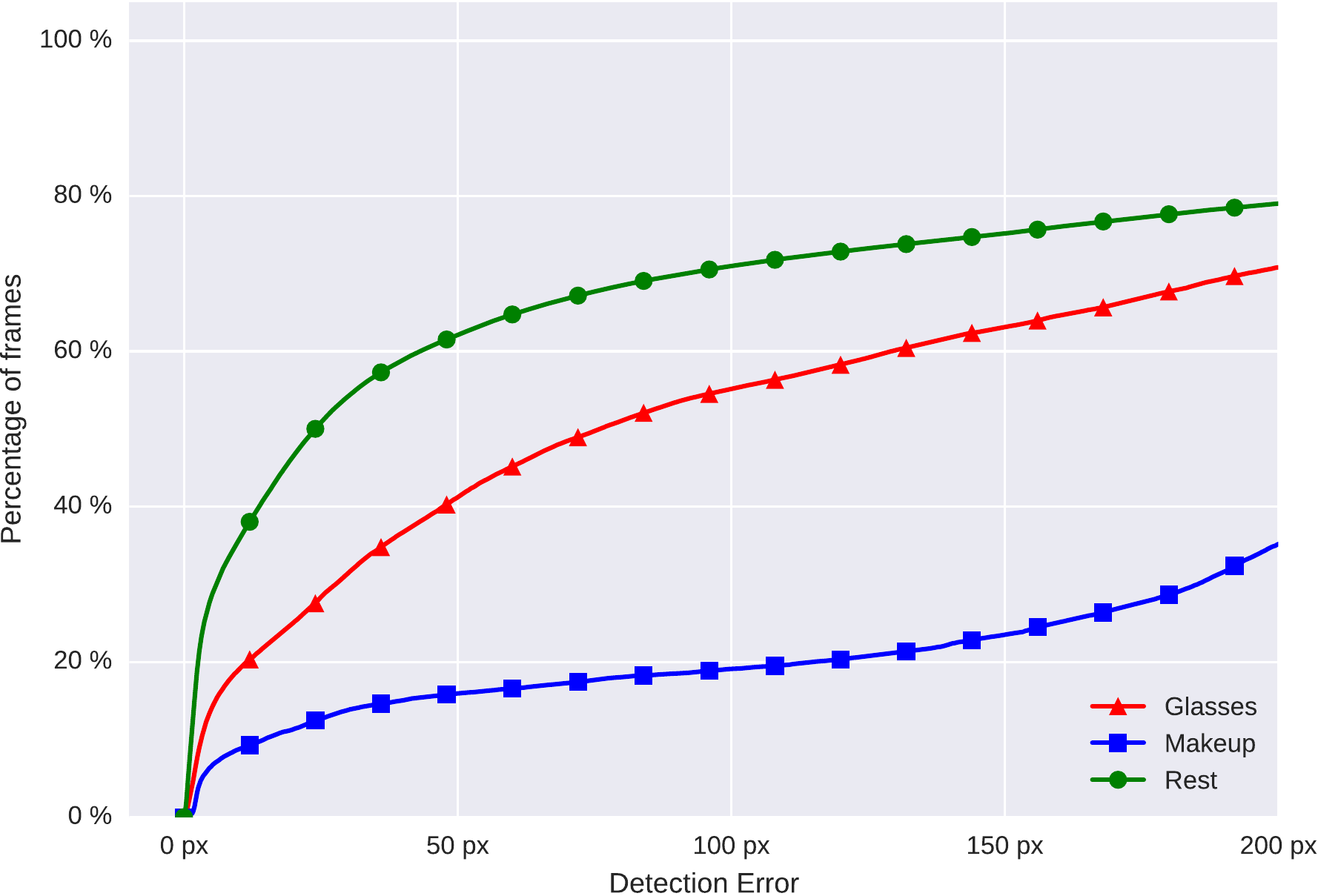}
    	\caption{Glasses and makeup evaluation}
        \label{fig:galsses_makeup_eva}
    \end{subfigure}
    \begin{subfigure}[t]{0.33\textwidth}
    	\includegraphics[width=\columnwidth]{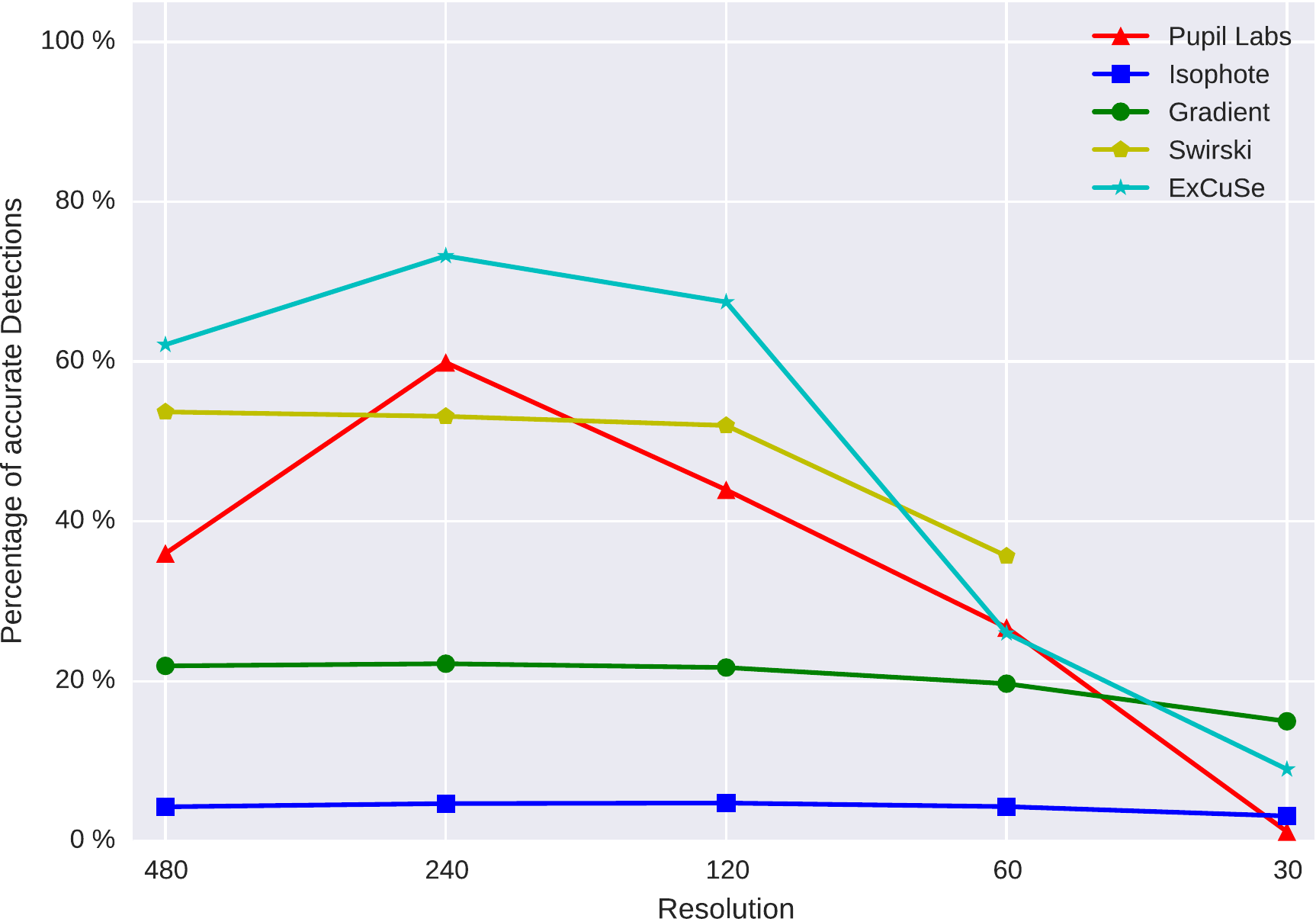}
    	\caption{Resolutions evaluation}
        \label{fig:resolution_eva}
    \end{subfigure}
\caption{Performance over different factors. Cumulative mean error distribution for indoor and outdoor videos of the 5 algorithms (a). The x-axis describes the detection error in pixels, while the y-axis describes the percentage of detections that had an error equal or lower to the corresponding x-value. A similar cumulated error distribution for the data that either include glasses, makeup or neither (b). Performance of each algorithm for images scaled to different resolutions (c). The x-axis states the height of the used resolution in pixels (ratio of 4:3 is fixed). The y-axis describes the percentage of detections with normalized error smaller than 0.02 of the corresponding resolution.}
\label{fig:indoorGlassesResolution}
\end{figure*}

Outdoor images are especially challenging for pupil detection algorithms, since the infrared portion of strong sunlight can create intense reflections and shadows on the pupil and iris (see also Figure \ref{fig:teaser} (e), (i) and (k)). Light falling directly into the camera lense can create additional reflections. Figure~\ref{fig:in_out_door_eva} shows the cumulative error distribution for the mean error of all algorithms for indoor and outdoor scenes. While on indoor scenes roughly 60\% of all detections had an error of 50px or lower, on outdoor scenes it is only about 50\%.

\subsubsection*{Glasses and Makeup}
For users with impaired eyes, the possibility to wear glasses along with the eye tracker is very important. However, glasses can cause intense reflections in the images and the pupil will often be partially occluded (see also Figure \ref{fig:teaser} (g) and (l)). The performance of the examined algorithms is significantly worse for participants wearing glasses compared to ones without glasses (see the Figure~\ref{fig:galsses_makeup_eva}).
According to our evaluation, makeup also greatly disturbs the performance of the examined algorithms, which is also visible in Figure~\ref{fig:galsses_makeup_eva}. One could expect this, since all algorithms either look for large black blobs or strong edges, which both could be also created by makeup. 

\subsubsection*{Resolution}
The examined algorithms have been designed for different systems working with different image resolutions. Namely the {\em Isophote} and {\em Gradient} detectors have been designed to work on low-resolution images while the others are usually for higher resolutions. In Figure~\ref{fig:resolution_eva}, we show the performance of each algorithm for different resolutions. The error is normalized by image width, and the percentage of detections with an error lower then 0.02 is shown. Parameters depending on the image size have been modified accordingly for all algorithms. The results for 30p of {\em Swirski} are missing because we couldn't get it to work on that resolution. 
It is important to note that in the implementations of the  {\em Gradient} and {\em Isophote} detector the input image was by default already downsampled to $80 \times 35$ pixels. Thus the performance for those algorithms remains constant, except for the smallest resolutions. As one can see the other algorithms all start to drop significantly in performance at some point while decreasing the resolution, until the performance becomes equal or worse to the former mentioned method.
Interestingly, the performances of {\em Swirski} and {\em ExCuSe} improved when downsampling from 480p to 240p. It indicates that 240p resolution is already enough for those methods, and higher resolution can harm the performance possibly due to increased image noise. 

%% file: 06_discussion.tex

\section{Discussion}
In this paper we presented a novel dataset for the development and evaluation of pupil detection algorithms. Our goal was to collect a comprehensive set of unconstrained high-quality recordings in realistic day-to-day environments and to go beyond the difficulties provided by other existing datasets. Also we evaluated the performance of state-of-the-art algorithms on our dataset. As the evaluation has shown, none of the examined algorithms performed well on all parts of the dataset. The detection accuracy in at least half of all cases was not sufficient to ensure a good eye tracking performance. This highlights the general difficulty of pupil detection in day-to-day environments and indicates the need to improve upon current algorithms. Further we were able to identify some of the key-challenges in those environments, which can give researchers an idea about what problems to focus on. Especially the presence of glasses and makeup could be shown to be a severe problem for current algorithms. Also the difficulty of performing on images recorded outdoors was highlighted in comparison to images recorded indoors. Further the influence of image resolution has been evaluated. While this identification of challenges is not yet complete, it highlights many open problems and can serve as a reference when developing new approaches. Given it's high quality, size and difficulty, our dataset serves as a good benchmark for evaluating new algorithms. The videos have been recorded in realistic day-to-day environments, however the actual viewing behaviour of the participant was controlled via a gaze target and is thus not natural. Given the videos high FPS, the development of tracking based algorithms can be considered.

%% file: 07_conclusion.tex

\section{Conclusion}

We presented labelled pupils in the wild (LPW), a novel dataset of eye region videos for the development and evaluation of pupil detection algorithms.
Our dataset includes 66 ground truth annotated, high-quality videos (130,856 frames) recorded from 22 participants in everyday locations at about 95 FPS; it is one order of magnitude larger than existing datasets.
Performance evaluations on the dataset demonstrated fundamental limitations of current pupil detection algorithms
and highlighted key challenges of head-mounted pupil detection due to lighting, image resolution, and vision aids.